\documentclass[runningheads,orivec]{llncs}
\usepackage[T1]{fontenc}
\usepackage{graphicx}
\newcommand{\approach}{MARS\xspace}

\newcommand{\ie}{i.\,e.,~}

\newcommand{\eg}{e.\,g.,~}

\newcommand{\sparqlreasoner}{SPARQLReasoner}

\renewcommand{\lambda}{\mathfrak{p}}
\usepackage{xspace}
\usepackage{amsmath}
\usepackage[colorlinks=false, hidelinks]{hyperref}
\usepackage[capitalize,nameinlink]{cleveref} 
\usepackage{color}
\usepackage{multirow}
\usepackage{tabularx}
\usepackage{booktabs}
\usepackage{todonotes}
\usepackage{csquotes}

\usepackage{amssymb}

\newcommand{\dataset}[1]{\texttt{#1}}
\newcommand{\tool}[1]{\textsc{#1}}

\usepackage[inline]{enumitem}
\usepackage{multirow}
\usepackage[table]{xcolor}
\usepackage{soul}
\usepackage{graphicx}
\usepackage{caption}
\usepackage[many]{tcolorbox}
\newtcolorbox{prompt}[1][]{
  fontupper=\scriptsize,
  fonttitle=\bfseries,
  colback=gray!10,
  colframe=gray!60,
  boxrule=0.5pt,
  arc=2mm,
  left=2mm,
  right=2mm,
  top=1mm,
  bottom=1mm
}
\usepackage{pdfpages}
\usepackage{array} 

\usepackage{ragged2e}              
\newcolumntype{Y}{>{\RaggedRight\arraybackslash}X}

\begin{document}
%
\title{MARS: Multi-hop Adaptive Retrieval and SPARQL Generation for KGQA}
\titlerunning{MARS: Multi-hop Adaptive Retrieval and SPARQL Generation for KGQA}
%

\author{Anonymous Author}
\author{Nikit Srivastava\orcidID{0009-0004-5164-4911} \and
Daniel Vollmers\orcidID{0000-0002-5324-4952} \and
Ren\'e Speck\orcidID{0009-0001-0151-5538}\and
Nikolaos Karalis\orcidID{0000-0002-0710-7180}\and \\
Hamada M. Zahera\orcidID{0000-0003-0215-1278}\and
Axel-Cyrille Ngonga Ngomo\orcidID{0000-0001-7112-3516}}
\authorrunning{Anonymous Author}
\authorrunning{N. Srivastava et al.}
%

\institute{Data Science Group (DICE), Heinz Nixdorf Institute, \\Department of Computer Science, Paderborn University, Germany\\
\email{\{nikit.srivastava,axel.ngonga\}@uni-paderborn.de}}

%
\maketitle              
%

%
%
%

\begin{abstract}

Large language models (LLMs) have demonstrated strong reasoning performance, but their tendency to hallucinate limits their reliability in knowledge-intensive tasks requiring up-to-date and grounded information.  
Combining knowledge graphs (KGs) with LLMs facilitates the use of explicit symbolic knowledge that can be continuously updated without costly fine-tuning, while benefiting from rapidly advancing LLM reasoning.
We propose MARS, a scalable knowledge graph question answering (KGQA) approach that requires no model fine-tuning.
Rather than relying on open-ended agentic exploration, MARS performs a structured retrieval procedure that links question entities to the KG and iteratively retrieves relevant next-hop information. 
At each step, MARS decides whether to continue graph traversal or to generate the final SPARQL query, allowing the model to adapt the retrieval depth to the question while keeping the overall pipeline more predictable than fully agentic approaches. We evaluate MARS on three established KGQA benchmarks across several LLMs and settings, including multilingual evaluation, and provide insights through ablation studies and error analysis.
Our approach achieves competitive performance relative to state-of-the-art methods while remaining efficient and scalable.
The evaluation results, code and resources are publicly available: \url{https://github.com/dice-group/mars-kgqa}.

\end{abstract}

\section{Introduction}\label{sec:intro}
Large language models (LLMs) have achieved strong performance across diverse language and knowledge-intensive tasks \cite{jiang2023mistral7b,gemmateam2025gemma3technicalreport,qwen3}, but remain prone to hallucinations \cite{10.1145/3703155,10.1145/3618260.3649777}, factual inconsistency~\cite{goodrich2019}, and multi-hop reasoning failures~\cite{krishna-etal-2025-fact}. 
While retrieval-augmented generation (RAG) \cite{krishna-etal-2025-fact,lewis2020} mitigates these challenges through external grounding, text-based RAG systems exhibit limitations with complex multi-hop reasoning as relational structure is only implicitly represented in textual contexts \cite{hu-etal-2025-grag,yang-etal-2024-large-language-models}. 
Incorporating symbolic knowledge, such as KGs, during retrieval preserves relational semantics and improves performance on complex multi-step questions \cite{hu-etal-2025-grag,10.1145/3777378}.
Fact‑retrieval approaches \cite{baek-etal-2023-direct,ICLR2024_10a6bdca} retrieve KG subgraphs according to their relevance to an input question and perform LLM reasoning to find an answer. 
These approaches become inefficient when query‑relevant subgraphs exceed the LLM’s context window \cite{li2025simple}.
For example, answering the question 
\textit{"How many people live in cities in the vicinity of the Nile?"}
requires structured aggregation over many entities and relations, which is difficult to perform reliably through direct reasoning over retrieved triples alone. 

In such cases, generating executable SPARQL queries provides a more scalable and interpretable alternative. 
However, systems fine-tuned on Question-SPARQL pairs \cite{savkin-etal-2024-deeppavlov,srivastava2024mst5multilingualquestion,vollmers_2024_ekaw} face challenges in SPARQL query generation \cite{walter2025graspgenericreasoningsparql} and are limited by training‑data coverage and diversity.  
More recent agent-oriented approaches leverage code-pretrained LLMs for SPARQL composition and graph exploration \cite{liu-etal-2024-spinach,walter2025graspgenericreasoningsparql}. 
These systems are vulnerable to inconsistent behavior due to their heavy reliance on LLMs, and thus, to hallucination at each step \cite{luo2025largelanguagemodelagent}. 
Graph-based RAG approaches \cite{hu-etal-2025-grag,10.1145/3777378} often retrieve and reason over graph neighborhoods or raw triples.
Such methods may require prior graph summarization and subgraph indexing, which introduce significant computational and storage costs in large-scale settings  \cite{10.1007/978-981-95-3061-8_1}.
Moreover, as the number of entities and relations in the KG grows, the candidate retrieval space can expand rapidly, making efficient and precise retrieval increasingly difficult \cite{10.1145/3703155,10.1145/3777378,xiang2026when}. 

To address these limitations, we propose \approach, a SPARQL‑based KGQA approach that combines pattern-based graph retrieval with context-augmented knowledge for knowledge-driven reasoning without requiring fine-tuning. 
\approach constrains the retrieved context by filtering and ranking \textit{triple patterns} rather than operating directly on subgraphs or raw triples. 
By combining retrieved pattern instances and typological schema information, \approach provides a KG‑grounded context based on a triple pattern set that enhances the LLM's capacity for SPARQL generation. When needed, it  expands this set by extracting new graph structures, thereby maintaining graph alignment throughout the multi-hop generation process.  
Evaluated across three benchmark datasets, \approach achieves state-of-the-art performance on the KGQA task without fine-tuning, consistently outperforming robust baselines.   
Notably, on the highly complex QALD-10 benchmark, with over 29\% of the queries requiring multiple hops, our approach outperforms agentic alternatives across all evaluated languages.
Furthermore, we address a critical reproducibility challenge in KGQA: benchmark datasets are often released without the specific KG version used during their creation \cite{10.1007/978-3-030-30796-7_5,perevalov2022qald}, or with versions lacking the coverage required to answer the queries. 
For instance, the Wikidata \cite{vrandevcic2014wikidata} truthy dump provided by QALD-10~\cite{usbeck2024qald} omits essential property statements and qualifiers for some of the benchmark queries. 
To ensure fair comparison and reproducibility for future work, we release the underlying Wikidata snapshot alongside the updated datasets and employ open-weight LLMs to ensure our experiments are completely reproducible.
Our work makes the following key contributions: 
\begin{enumerate}
\item A KGQA approach that curtails 
LLM context size with pattern‑based graph retrieval and iteratively incorporates next‑hop facts into LLM reasoning, 
\item A hybrid multilingual setup that uses translation alongside the native question, covering 10 languages, including four low‑resource languages, across three KGQA benchmark datasets,
\item An in-depth ablation study on features and configurations, 
and an error analysis of our approach and results,
\item Reproducibility and open access for all results
, code, and resources.
\end{enumerate}

\noindent
In the following section, we review related work before we describe our approach in detail in Section \ref{sec:approach}.
Section \ref{sec:exp-setup} outlines our experimental setup, followed by our evaluation and result discussion alongside error analysis in Section \ref{sec:results}. 
We cover our ablation study in detail in Section \ref{sec:ablation}. 
We conclude our paper in Section \ref{sec:Conclusion}. 

\section{Related Work}\label{sec:relwork}
\subsection{Knowledge Graph Question Answering}
Recent work explores the integration of KGs in LLMs for KGQA~\cite{Khorashadizadeh24,pan2024unifying,ICLR2024_10a6bdca}. 
Although LLMs show promising results in reasoning~\cite{Tan23}, integrating structured data into KGQA remains challenging \cite{walter2025graspgenericreasoningsparql}. 
%
Context-augmented approaches~\cite{Baek2023,jiang-etal-2023-structgpt,Meyer24,mialon2023augmented} with autoregressive models demonstrate improved performance in zero-shot and few-shot scenarios. 
Relevant evidence collected from structured data with an iterative reading-then-reasoning approach in zero-shot and few-shot scenarios achieved promising results with LLMs~\cite{jiang-etal-2023-structgpt}.
\tool{ChatGPT}~\cite{ouyang2022traininglanguagemodelsfollow} and \tool{Llama}‑based models~\cite{grattafiori2024llama3herdmodels} generate syntactically correct SPARQL queries and reduce computational cost (\eg on DBpedia \cite{10.5555/1785162.1785216})
 but remain below optimal performance on complex Wikidata queries, non‑readable identifiers and non‑English inputs~\cite{mecharnia2025performance,Meyer24}.
Data augmentation using KG axioms improves coverage and performance~\cite{rangel2024sparql}, while models like Mistral~\cite{jiang2023mistral7b} and GPT-3.5 outperform T5~\cite{t5} on \dataset{LC-QuAD1.0} and \dataset{QALD-9}~\cite{Longwell24}. 
Recent advances include RAG-based federated SPARQL query generation~\cite{emonet24}, dynamic few-shot learning~\cite{dabramo-etal-2025-investigating}, modular designs with RAG layers~\cite{pan2025}, and LLM SPARQL query generation under different prompting conditions~\cite{gashkov2025sparqlquerygenerationllms}. 
Prompt-based methods reduce labeled data dependency but are sensitive to prompt design~\cite{kovriguina2023sparqlgen}.
Agentic frameworks such as \tool{mKGQAgent}~\cite{perevalov2025texttosparqlgoesenglishmultilingual}, \tool{Agentict$^2$s}~\cite{zhao2025agentic}, \tool{SPINACH}~\cite{liu-etal-2024-spinach} and \tool{GRASP}~\cite{walter2025graspgenericreasoningsparql}  are vulnerable to error propagation due to their heavy reliance on LLMs with  hallucinations. 
%
However, many existing frameworks perform a lossy and unidirectional conversion of non-English queries into English prior to retrieval~\cite{ICLR2024_10a6bdca}, largely because downstream KGQA components for entity linking are predominantly optimized for English~\cite{10.1145/3587259.3627567}. 
While machine translation offers a viable approach~\cite{10.1145/3485447.3511940,10.1145/3587259.3627567}, processing queries directly in the user's native language is preferable for preserving the original semantic intent~\cite{srivastava2024mst5multilingualquestion}.

\approach employs a hybrid strategy that combines machine translation and entity-type augmentation~\cite{vollmers2025eleval,vollmers-etal-2025-contextual} with the original native-language text. This allows \approach to preserve the query's original structure for semantic pattern matching while simultaneously leveraging the advantages of machine translation.

\subsection{Reasoning in Large Language Models}
LLMs have demonstrated strong reasoning abilities over provided contexts \cite{dasgupta2024languagemodelshumanlikecontent,suzgun2022challengingbigbenchtaskschainofthought,wei2022emergentabilitieslargelanguage}. 
Notable examples such as  
Gemini \cite{comanici2025gemini25pushingfrontier}, Claude \cite{TheC3}, and DeepSeek‑R1 \cite{deepseekai2025deepseekr1incentivizingreasoningcapability} achieve state‑of‑the‑art performance on a wide variety of reasoning tasks \cite{ke2025surveyfrontiersllmreasoning}.
LLM reasoning can be classified as implicit or explicit. 
Implicit reasoning produces the final answer directly and is computationally efficient, whereas explicit reasoning first generates an intermediate ``thinking'' context before generating the answer, which improves transparency and explainability \cite{li2025implicitreasoninglargelanguage}.
Leading implicit-reasoning systems include Gemma~\cite{gemmateam2025gemma3technicalreport}, Qwen~\cite{qwen3}, and Mistral. 
Among explicit‑reasoning approaches, state‑of‑the‑art models include GPT‑OSS, DeepSeek‑R1, Llama, and GLM‑4.5 \cite{openai2025gptoss120bgptoss20bmodel,5team2025glm45agenticreasoningcoding}.

Rather than requiring the LLM to act as an autonomous agent that selects tools at runtime ~\cite{walter2025graspgenericreasoningsparql}, which can increase hallucinations, \approach{} supports a deterministic, pattern-guided abstraction loop that allows the model to concentrate solely on path pruning and the final query generation. 
Moreover, we focus on open‑weight models for reproducibility.

\section{Our Approach} \label{sec:approach}

\subsection{Task Definition and Preliminaries}\label{sec:pre}
To address the problem of answering natural‑language questions over RDF knowledge graphs, we first formalize the inputs and objectives of our approach.

\subsubsection{Knowledge Graph Question Answering} 
Given a natural language question $q_{text}$ and an RDF knowledge graph
$\mathcal{G} \subseteq \mathcal{E} \times \mathcal{P} \times (\mathcal{E} \ \cup\ \mathcal{L} \ \cup\ \mathcal{C})$
that consists of triples $(s, p, o) \in \mathcal{G}$, where 
$\mathcal{E}$ denotes a set of entities, 
$\mathcal{P}$ a set of predicates (\ie relations), 
$\mathcal{L}$ a set of literals, and  
$\mathcal{C}$ a set of classes, 
the main objective of our approach is to answer the question $q_{text}$ based on $\mathcal{G}$ by expressing it as a SPARQL query $q_{sparql}$, the standard query language for RDF KGs:
\begin{align}
q_{sparql} &= \mathrm{KGQA}(q_{text}, \mathcal{G}) \label{eq:sparql_gen}.
\end{align}
Executing this generated query against $\mathcal{G}$ yields the answer set $\mathcal{A}$:
\begin{align}
\mathcal{A} &= \mathrm{Answer}(q_{sparql}, \mathcal{G}).\label{eq:approach_answer}
\end{align}
This answer set $\mathcal{A}$ represents the solution to question $q_{text}$ found within the knowledge graph $\mathcal{G}$.

\subsubsection{Knowledge Graph Schema}
Our system operates on a given KG, thus we rely on a given Terminological Box (TBox) that defines the schema of the KG, including axioms $\langle . \rangle$ that specify domain and range constraints for predicates.
We denote the domain  and range classes of a predicate $p$ with $\text{domain}(p)=\{ c \in \mathcal{C} | \langle p, \text{rdfs:domain}, c \rangle \in \text{TBox}\}$ and $\text{range}(p)=\{ c \in \mathcal{C} | \langle p, \text{rdfs:range}, c \rangle \in \text{TBox}\}$. 
Furthermore, let each term $t \in  \mathcal{T} =  \mathcal{E} \cup \mathcal{P} \cup \mathcal{C}$ 
assigned a label $\ell:\mathcal{T}  \to  \mathcal{L}$ with $\langle t , \text{rdfs:label}, l \rangle \in TBox$, 
where $l \in \mathcal{L}$ is a literal value, \ie human-readable name or description. We denote a label $l$ of a term $t$ with $\ell(t)$.

\subsection{Pipeline Overview} 
We propose a three-stage pipeline for multilingual KGQA: entity extraction, pattern extraction,\footnote{Our terminology for patterns is synonymous with triple patterns in SPARQL: \url{https://www.w3.org/TR/2013/REC-sparql11-query-20130321/\#sparqlTriplePatterns}} and a final stage that either generates the output query or expands the context.
In the latter case, the newly retrieved resources enrich the context, and the pattern extraction stage is re-invoked to capture next-hop information. This loop continues until the query is fully grounded, at which point the final SPARQL query is generated.

\begin{figure}[t!]
\centering
    \includegraphics[width=0.9\textwidth]{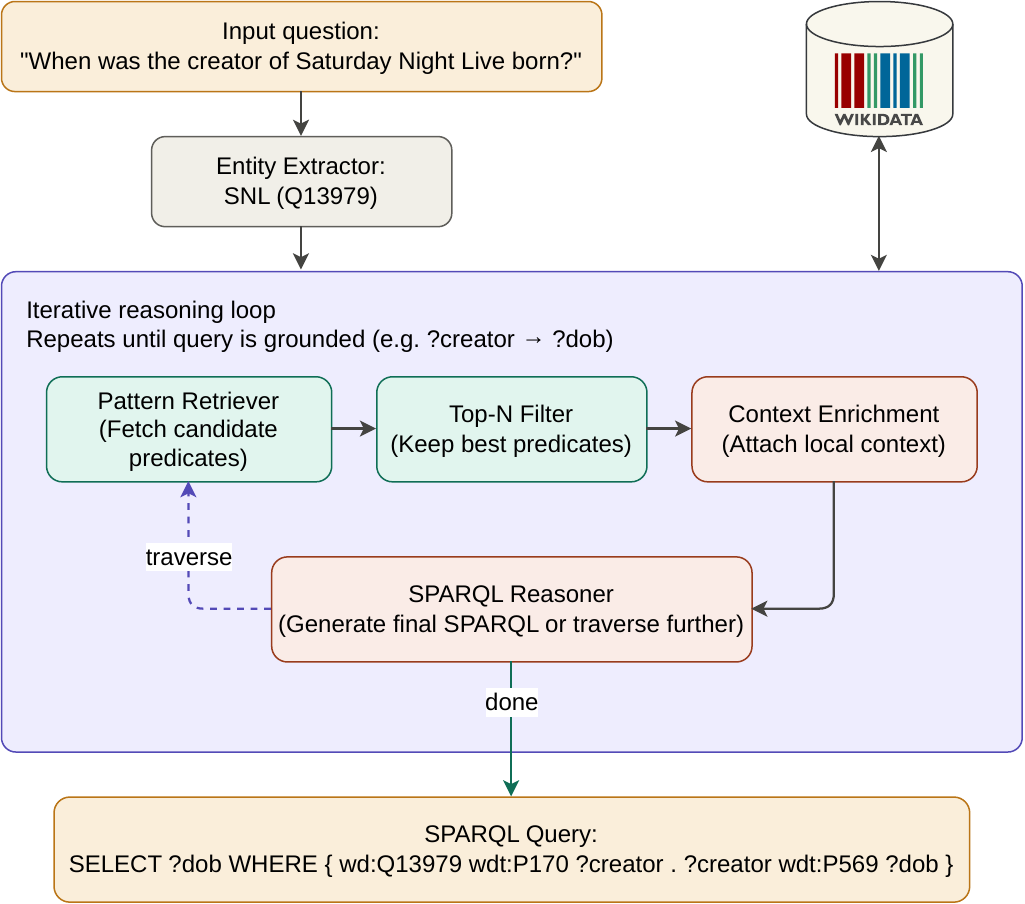}
    \caption{\approach pipeline for an example question. }
    \label{fig:mars-overview}    
\end{figure}

An overview of our \approach pipeline is depicted in Figure \ref{fig:mars-overview}, using the example question: ``\textit{When was the creator of Saturday Night Live born?}'' (from the \dataset{QALD-10} dataset). The process begins with entity extraction and moves into the iterative reasoning loop (highlighted in light purple), where the system retrieves patterns and enriches context until the query is grounded. The final output is a generated SPARQL query, with the process flow indicated by the arrows.\footnote{Detailed image: \href{https://github.com/dice-group/mars-kgqa/blob/master/data_dir/figures/mars-kgqa-overview.png}{mars-kgqa/data\_dir/figures/mars-kgqa-overview.png}} We provide details for each step of our approach in the following. 

\subsection{Entity Linking and Text Augmentation}\label{subsec:ent_ext}

We first identify entities $E \subseteq \mathcal{E}$ in $q_{text}$ that exist within $\mathcal{G}$ using the following operation:
\begin{equation}
E = \mathrm{ExtractEntities}(q_{text}, \mathcal{G}).
\label{eq:approach_el}
\end{equation}
$\mathrm{ExtractEntities}(\cdot)$ is implemented using an entity linking pipeline based on \break\tool{GRASP}~\cite{walter2025graspgenericreasoningsparql}.\footnote{Our modified version: \href{https://github.com/dice-group/grasp_el/blob/main/ANNOTATION_PIPELINE.md}{grasp\_el/ANNOTATION\_PIPELINE.md}} To further support the graph traversal, we adopt a text augmentation strategy~\cite{vollmers2025eleval,vollmers-etal-2025-contextual} that utilizes an instruction-tuned LLM to generate $q^{\textsc{aug}}_{text}$, which enriches the query context by incorporating predicted entity type labels. 
To support non-English queries in both of these steps, we employ a multilingual LLM \cite{gemmateam2025gemma3technicalreport} for translation.

\subsection{Pattern Retrieval and Filtering} \label{sec:pe}

The aim of this step is to retrieve a set of patterns 
to extract first-hop information from $\mathcal{G}$ about the extracted entities $E$, thus about the question $q_{text}$.

Using a set of patterns, instead of full triples, significantly reduces the amount of context retrieved from the graph. 
Each extracted pattern consists of a predicate, an entity, and a placeholder variable. 
The entity is positioned as either subject or object depending on the edge's direction.
If the entity functions as the subject, the placeholder serves as the object, and vice versa.
We impart this functionality by defining a set of patterns:

\begin{equation}
\mathfrak{P} = \left\{ (s, p, \_) \mid s \in E, (s, p, o) \in \mathcal{G}\, \right\}  \cup \left\{ (\_, p, o) \mid o \in E, (s, p, o) \in \mathcal{G}\, \right\}. 
\label{eq_pattern}
\end{equation}
\noindent
The retrieved patterns are then verbalized ($\oplus$ denotes string concatenation): 
\begin{equation}
\begin{aligned}
\mathsf{V}_{\mathfrak P}
&= \{\, \mathsf{v}(\lambda) \mid \lambda \in \mathfrak P \,\}, \text{with} \\
\mathsf{v}(s,p,o)
&=
\begin{cases}
\operatorname{\ell}(s) \oplus \operatorname{\ell}(p) \oplus \texttt{"?object"} & \text{if } o = \_,\\[1mm]
\texttt{"?subject"} \oplus \operatorname{\ell}(p) \oplus \operatorname{\ell}(o) & \text{if } s = \_,\\[1mm]
\end{cases} . 
\end{aligned}
\label{eq:verbalize_patterns}
\end{equation}
We then convert each verbalized pattern in $\mathsf{V}_{\mathfrak P}$ and the input question $q_{\text{text}}$ (or $q^{\textsc{aug}}_{text}$ depending upon the configuration, see Section \ref{sec:ablation}) into $d$-dimensional dense embedding vectors $\Phi_{\mathfrak P}=\{ \phi(\mathsf{v}) |\mathsf{v} \in \mathsf{V}_{\mathfrak P}  \}$ and $\phi(q_{\text{text}})$ with the text embedding function $\phi: \Sigma^* \to \mathbb{R}^d$. 
These vectors are then ranked by their semantic similarity, \ie  dot product, to the vector of the input question:
\begin{equation}
\begin{aligned}
\mathfrak{S}_{\mathfrak P} &= \mathrm{SemanticSimilarity}(\Phi_{\mathfrak P}, \phi(q_{\text{text}})).
\end{aligned}
\label{eq:semantic_similarity}
\end{equation}
Finally, we select the top-$N$ 
patterns with $\textsc{topn} \in \mathbb{N}^+$:
\begin{equation}
\begin{aligned}
\mathfrak{P}_\text{top} &= \mathrm{SelectTopN}({\mathfrak P}, \mathfrak{S}_{\mathfrak P}, \textsc{topn} ).
\end{aligned}
\label{eq:select_top_n}
\end{equation}
\subsection{Enrichment and Multi-Hop Reasoning}

\noindent
Once the top-$N$ patterns are obtained, 
we extract features that capture information critical for pattern understanding.
These features include retrieving instance values, frequencies, and typological information based on the pattern's properties. 
Let $\lambda$ be an example pattern drawn from the set $\mathfrak{P}_\text{top}$. 
We define these features in the following.

\subsubsection{Pattern Instances}
We retrieve all instances associated with variables in $\lambda$:
\begin{equation}
\begin{aligned}
\operatorname{I}_{\lambda}
   = \{\, o \mid \lambda = (s,p,\_),(s,p,o)\in\mathcal{G}, o \in \mathcal{E}  \} 
\cup \{\, s \mid \lambda = (\_,p,o),(s,p,o)\in\mathcal{G}\,\} . \\[2mm]
\end{aligned}
\label{eq:pattern_instances_all}
\end{equation}
We operate based on the pattern structure. 
The first case, $\lambda = (s,p,\_)$, extracts all objects ($o$) related to subject $s$ and predicate $p$. 
The second case, $\lambda = (\_,p,o)$, extracts all subjects ($s$) related to predicate $p$ and object $o$.


\subsubsection{Typological Information}
A verbalization function $\mathsf{t}(p)$ for typological information of the pattern $\lambda$ based on the domain and range of its predicate $p$ within the pattern: 
$\mathsf{t}(p)=\texttt{"domain:"} \bigoplus_{c \in \text{domain}(p)} \ell(c) \texttt{", range:"} \bigoplus_{c \in \text{range}(p)} \ell(c)$.





\subsubsection{Enriched Verbalization}
We then generate an enriched verbalization incorporating these features that can be defined as:

\begin{equation}
\begin{aligned}
\mathsf{V}^{\text{enriched}}_{\mathfrak{P}_\text{top}}
&= \Bigl\{\,\mathsf{v}(\lambda)
        \;\oplus\;
        n 
        \;\oplus\; L 
        \;\oplus\;
        \operatorname{t}(p)
        \;\Big|\;
        \lambda=(s,p,o) \in\mathfrak{P}_{\text{top}}
    \Bigr\}, \\[2mm]
\end{aligned}
\label{eq:enrich_verbalization_compact}
\end{equation}
with $n=|\operatorname{I}_{\lambda}|$, the pattern frequency by simply counting the number of instances, and with 
$L = \bigoplus_{j \in I_i} \ell(j)$,
the concatenation of instance labels, where we focus on instance subsets  to reduce computational cost:  
$\operatorname{I}_i \subseteq\operatorname{I}_{\lambda}$,
 with $|\operatorname{I}_i|=i\in \mathbb{N}^+$.

\subsubsection{SPARQL Query Generation and Reasoning}
This enriched verbalization effectively combines schema information with concrete instances, revealing the question-relevant context of the graph. 
The enriched verbalization and the original question are then passed to a SPARQL Reasoner\footnote{\href{https://github.com/dice-group/mars-kgqa/blob/master/supplementary_material/appendix.md\#sparqlreasoner-prompt}{mars-kgqa/supplementary\_material/appendix.md\#sparqlreasoner-prompt}}, i.e., an LLM, preferably one with strong reasoning and coding capabilities:

\begin{equation}
\begin{aligned}
&\operatorname{\sparqlreasoner}\Bigl(q_{\text{text}},\,
\mathsf{V}^{\text{enriched}}_{\mathfrak{P}_{\text{top}}}\Bigr)\;
=\;
\begin{cases}
\displaystyle
q_{\text{sparql}}
& \text{if } 
\operatorname{qgen}\Bigl(q_{\text{text}},\,
\mathsf{V}^{\text{enriched}}_{\mathfrak{P}_{\text{top}}}\Bigr), \\[2.2ex]
\displaystyle
\mathfrak{P}_{\text{sel}}
& \text{otherwise}.
\end{cases}
\end{aligned}
\label{eq:sparql_reasoner_conditional}
\end{equation}

\noindent
The reasoner determines whether it has sufficient context to generate the final SPARQL query $q_{\text{sparql}}$, which we represent as a boolean function $\operatorname{qgen}(\cdot)$. 
If the reasoner cannot generate a query, it selects the specific patterns $\mathfrak{P}_{\text{sel}} \subseteq \mathfrak{P}_{\text{top}}$ that are then expanded in the next round of pattern extraction.

We define the expanded node set $E_{\text{vis}} \subseteq  \mathcal{E} $ from an entity set $E$ and top patterns $\mathfrak{P}_{\text{top}}$ with $E_{\text{vis}} 
=  \bigcup_{ \mathfrak{p} \in \mathfrak{P}_{\text{top}}}^{}{ \operatorname{I}_{\mathfrak{p}}} \cup E
$.
The non-visited nodes are then extracted as 
$E_{\text{next}} = E_{\text{sel}} \setminus E_{\text{vis}}$,
where $E_{\text{sel}} = \{ s | (s,p,\_) \in \mathfrak{P}_{\text{sel}} \}  \cup \{ o | (\_,p,o) \in \mathfrak{P}_{\text{sel}}\}$ is a set of all entities inside $\mathfrak{P}_{\text{sel}}$. 
Then, we set $E:= E_{\text{next}}$ and repeat the pattern extraction process 
to 
extract next-hop information  $\mathfrak{P}^{next}$ with \cref{eq_pattern}. 
Afterward, we select the top-$N$ from these $\mathfrak{P}_{top}^{next}$ with \cref{eq:select_top_n}.  

Finally, we repeat the process in this section by adding these new patterns to the accumulated list and re-attempting SPARQL query generation with \cref{eq:sparql_reasoner_conditional}:
\begin{equation}
\begin{aligned}
\mathfrak{P}_{top} = \mathfrak{P}_{top} \cup \mathfrak{P}_{top}^{next},
\end{aligned}
\label{eq:repeat_loop}
\end{equation}
until a SPARQL query is successfully generated or a pre-defined multi-hop limit, \ie  \textsc{mhop}, is reached. 

The final values for variables $d$ and $i$, along with other configuration details, are presented in Section \ref{subsec:mars_config}. Further investigation into optimal values for \textsc{topn} and \textsc{mhop} is conducted as part of our ablation study in Section~\ref{sec:ablation}.

Our prompting format draws inspiration from earlier retrieval and reasoning approaches \cite{ICLR2024_10a6bdca,walter2025graspgenericreasoningsparql}, yet it has been adapted to suit our specific use case. While the instructions are written in English, the question itself is always retained in its original language. 

\section{Experimental Setup} \label{sec:exp-setup}

\subsection{Datasets} \label{subsec:datasets}

We evaluate our approach on three widely used Wikidata-based KGQA benchmark datasets, each addressing distinct evaluation dimensions.  
\dataset{LCQuAD2.0}~\cite{10.1007/978-3-030-30796-7_5} provides a large-scale English-only benchmark (30k training, 6k test), establishing baseline performance for standard queries. \dataset{QALD-9-plus}~\cite{perevalov2022qald} introduces multilingual evaluation (9 languages, 371 training and 136 test) to test cross-lingual capabilities across diverse language families. \dataset{QALD-10}~\cite{usbeck2024qald} challenges systems with 394 complex test queries (4 languages) featuring advanced SPARQL constructs (aggregations, subqueries), testing reasoning robustness beyond simple fact retrieval. The questions in \dataset{LCQuAD2.0} are produced from system‑generated templates, whereas the \dataset{QALD} benchmarks consist of queries that reflect real-world information needs of proficient English users and are translated into the other languages by native speakers. Collectively, these datasets enable comprehensive assessment of scalability, multilingual adaptability, and complex reasoning capabilities. 
Additionally, we share our analysis on the multi-hop queries on the test splits of these datasets in Table \ref{tab:mhop_analysis}.
\begin{table}
\caption{Proportion of multi-hop queries across datasets.}
\label{tab:mhop_analysis}
\centering
\resizebox{0.7\textwidth}{!}{%
\rowcolors{2}{white}{gray!20}
\setlength{\tabcolsep}{5pt}
\begin{tabular}{l r r r}
\toprule
\textbf{Dataset} & \textbf{1-hop} & \textbf{2-hop} & $\geq\,\text{\textbf{3-hop}}$  \\
\midrule
\dataset{QALD-9plus} & 97 (76.38\%) & 29 (22.83\%) & 1 (0.79\%)  \\
\dataset{QALD-10} & 268 (70.16\%) & 97 (25.39\%) & 17 (4.45\%)  \\
\dataset{LC-QuAD2.0} & 3053 (66.03\%) & 1571 (33.97\%) & 0  \\
\bottomrule
\end{tabular}
}%
\end{table}
%
%
For consistent evaluation, we use a specific Wikidata-dump\footnote{\href{https://github.com/dice-group/mars-kgqa/blob/master/supplementary_material/resources.md\#datasets}{mars-kgqa/supplementary\_material/resources.md\#datasets}} hosted locally with \tool{Tentris} \cite{10.1007/978-3-030-62419-4_4}. We chose \tool{Tentris} due to its faster query response times and stable scaling for queries requiring extensive join operations.
We serve this KG ($\approx$11B triples) as an HTTP endpoint for SPARQL query evaluation. To ensure fair comparison, we filtered reference queries yielding empty/failing results, resulting in test sets of 4624 (\dataset{LCQuAD2.0}), 127 (\dataset{QALD-9-plus}) and 382 (\dataset{QALD10}) pairs.

\subsection{Configuration and Hardware} \label{subsec:mars_config}
For the text embedding function $\phi$, we use a trained sparse mixture-of-experts text embedding model 
\cite{nussbaum2025trainingsparsemixtureexperts} with $d=768$ dimensions. 
For the semantic similarity, we applied the dot product, a standard and widely used NLP metric. 
For SPARQL query generation and reasoning, we limit retrieved instances to $i=10$ per pattern. 

To promote open-source development, we exclusively utilize open models and have tested \approach using several state-of-the-art open-weight LLMs\footnote{\href{https://github.com/dice-group/mars-kgqa/blob/master/supplementary_material/appendix.md\#models-tested-with-mars}{mars-kgqa/supplementary\_material/appendix.md\#models-tested-with-mars}} in our experiments. We utilize compute resources with 256 gigabytes of memory, 16 CPU cores of an AMD EPYC 7763 processor, and 4 Nvidia A100 GPU with 40GB of memory for hosting and experimenting with these LLMs.
\subsection{Baselines} \label{subsec:baselines}
We compare our approach against four SPARQL query generation baselines spanning diverse paradigms:
\tool{GRASP}
\footnote{We follow the original paper's configuration, but replace GPT-4.1 with the open-weight \tool{GPT-OSS} to align with this work's focus on open-weight models. The two models achieve broadly comparable performance on general benchmarks (\url{https://artificialanalysis.ai/models/comparisons/gpt-oss-120b-vs-gpt-4-1}).}
\cite{walter2025graspgenericreasoningsparql}, the current state-of-the-art, an agentic system that iteratively builds and executes SPARQL queries using a set of predefined tools;
\tool{DeepPavlov} \cite{burtsev-etal-2018-deeppavlov,savkin-etal-2024-deeppavlov}, a chat-based system combining templates with fine-tuned models for query prediction, entity recognition, and path ranking;
\tool{UniQ-Gen} \cite{vollmers_2024_ekaw}, a recent T5-based pipeline integrating entity linking, relation extraction, and query construction, designed to generalize across multiple knowledge graphs;
and \tool{MST5} \cite{srivastava2024mst5multilingualquestion}, a multilingual approach based on a fine-tuned mT5 that augments input questions with linguistic and entity information.
For systems other than \tool{MST5}, we use translated questions following prior work \cite{10.1145/3485447.3511940,10.1145/3587259.3627567}.
Together, these baselines combine state-of-the-art fine-tuned models with newer agentic systems.
\subsection{Metrics}
We evaluate  the performance of our system using four standard key performance indicators. 
\textbf{Precision} and  \textbf{Recall}
were calculated to compare system-generated answers against ground-truth results. Precision quantifies the correctness of answers retrieved by the generated SPARQL query against the ground truth. Recall measures the system’s coverage of all valid answers present in the knowledge graph. The F1-score balances precision and recall as a harmonic mean, reflecting overall accuracy in answer retrieval through the generated SPARQL query. In the context of this paper, we evaluate only with macro‑averaged metrics to avoid bias caused by class imbalance.
\textbf{Macro F1} 
combines macro precision and macro recall across test questions to assess SPARQL query's ability to retrieve the correct answer-set. 
\textbf{Macro F1 QALD}\footnote{\url{https://github.com/ag-sc/QALD}} is an adaption of Macro F1 which only considers samples where the QA-System predicted an answer for the metric computation.
All of the mentioned metrics are computed with the help of \tool{GERBIL-QA}\footnote{Note that the GERBIL-QA scoring implementation was updated in 2024 and 2026. Results may differ from earlier reports. 
\url{https://dice-research.org/GerbilQA}} \cite{DBLP:journals/semweb/UsbeckRHCHNDU19}.
\section{Results and Discussion}\label{sec:results}
\subsection{Performance Comparison}\label{sec:perf_comp}

We evaluate \approach on three benchmark datasets and report full details, including all key performance
indicators,  in the Appendix.\footnote{\href{https://github.com/dice-group/mars-kgqa/blob/master/supplementary_material/appendix.md\#performance-tables}{mars-kgqa/supplementary\_material/appendix.md\#performance-tables}} Table~\ref{tab:f1_combined} shows Macro F1 scores per dataset and language. \approach produces competitive or top scores in each setting, and across the three datasets it beats most baselines and outperforms \tool{GRASP} in 8 of 14 head-to-head comparisons. The weaker numbers of \tool{DeepPavlov}, \tool{MST5}, and \tool{UniQ-Gen} suggest that fine-tuning on Question-SPARQL pairs gives limited returns on heterogeneous benchmarks such as \dataset{QALD}. 
The performance gap between \approach and \tool{GRASP} indicates that, at least in our multi-hop question setting, a fixed pipeline can be more effective than an agent that selects tools dynamically at runtime. 

\begin{table}[htbp]
\caption{Macro F1 scores (in [\%]) of all systems across datasets.}
\label{tab:f1_combined}
\centering
\resizebox{\textwidth}{!}{%
\setlength{\tabcolsep}{3pt}
\begin{tabular}{l l c c c c | c}
\toprule
\textbf{Dataset} & \textbf{Lang.} & \textbf{\tool{DeepPavlov}} & \textbf{\tool{MST5}} & \textbf{\tool{UniQ-Gen}} & \textbf{\tool{GRASP}} & \textbf{\tool{\approach}} \\
\midrule
\rowcolor{gray!20}
\cellcolor{white} & ba & \href{https://gerbil-qa.aksw.org/gerbil/experiment?id=202605020007}{15.29} & \href{https://gerbil-qa.aksw.org/gerbil/experiment?id=202604160019}{7.20} & \href{https://gerbil-qa.aksw.org/gerbil/experiment?id=202605050108}{26.44} & \href{https://gerbil-qa.aksw.org/gerbil/experiment?id=202604270015}{26.18} & \href{https://gerbil-qa.aksw.org/gerbil/experiment?id=202605060002}{\textbf{36.32}} \\
  & be & \href{https://gerbil-qa.aksw.org/gerbil/experiment?id=202605020016}{21.42} & \href{https://gerbil-qa.aksw.org/gerbil/experiment?id=202604160016}{15.93} & \href{https://gerbil-qa.aksw.org/gerbil/experiment?id=202605050104}{44.57} & \href{https://gerbil-qa.aksw.org/gerbil/experiment?id=202604270022}{\textbf{52.14}} & \href{https://gerbil-qa.aksw.org/gerbil/experiment?id=202605060012}{49.56} \\
\rowcolor{gray!20}
\cellcolor{white} & de & \href{https://gerbil-qa.aksw.org/gerbil/experiment?id=202605020013}{21.74} & \href{https://gerbil-qa.aksw.org/gerbil/experiment?id=202604160010}{24.41} & \href{https://gerbil-qa.aksw.org/gerbil/experiment?id=202605050100}{43.78} & \href{https://gerbil-qa.aksw.org/gerbil/experiment?id=202604270009}{\textbf{57.43}} & \href{https://gerbil-qa.aksw.org/gerbil/experiment?id=202605060003}{56.51} \\
  & en & \href{https://gerbil-qa.aksw.org/gerbil/experiment?id=202605020008}{22.14} & \href{https://gerbil-qa.aksw.org/gerbil/experiment?id=202604160017}{25.20} & \href{https://gerbil-qa.aksw.org/gerbil/experiment?id=202605050105}{50.06} & \href{https://gerbil-qa.aksw.org/gerbil/experiment?id=202604270013}{\textbf{65.47}} & \href{https://gerbil-qa.aksw.org/gerbil/experiment?id=202605060004}{58.21} \\
\rowcolor{gray!20}
\cellcolor{white} & es & \href{https://gerbil-qa.aksw.org/gerbil/experiment?id=202605020009}{24.60} & \href{https://gerbil-qa.aksw.org/gerbil/experiment?id=202604160012}{22.84} & \href{https://gerbil-qa.aksw.org/gerbil/experiment?id=202605050101}{48.14} & \href{https://gerbil-qa.aksw.org/gerbil/experiment?id=202604270012}{\textbf{59.65}} & \href{https://gerbil-qa.aksw.org/gerbil/experiment?id=202605060006}{57.82} \\
  & fr & \href{https://gerbil-qa.aksw.org/gerbil/experiment?id=202605020014}{3.41} & \href{https://gerbil-qa.aksw.org/gerbil/experiment?id=202604160014}{5.25} & \href{https://gerbil-qa.aksw.org/gerbil/experiment?id=202605050102}{8.66} & \href{https://gerbil-qa.aksw.org/gerbil/experiment?id=202604270014}{12.38} & \href{https://gerbil-qa.aksw.org/gerbil/experiment?id=202605060001}{\textbf{71.93}} \\
\rowcolor{gray!20}
\cellcolor{white} & hy & \href{https://gerbil-qa.aksw.org/gerbil/experiment?id=202605020015}{2.62} & - & \href{https://gerbil-qa.aksw.org/gerbil/experiment?id=202605050106}{5.51} & \href{https://gerbil-qa.aksw.org/gerbil/experiment?id=202604270017}{10.14} & \href{https://gerbil-qa.aksw.org/gerbil/experiment?id=202605060000}{\textbf{65.96}} \\
  & ru & \href{https://gerbil-qa.aksw.org/gerbil/experiment?id=202605020017}{20.40} & \href{https://gerbil-qa.aksw.org/gerbil/experiment?id=202604160015}{22.84} & \href{https://gerbil-qa.aksw.org/gerbil/experiment?id=202605050103}{41.73} & \href{https://gerbil-qa.aksw.org/gerbil/experiment?id=202604270023}{\textbf{55.44}} & \href{https://gerbil-qa.aksw.org/gerbil/experiment?id=202605060007}{51.77} \\
\rowcolor{gray!20}
\cellcolor{white}\multirow{-9}{*}{\dataset{QALD-9-plus}} & uk & \href{https://gerbil-qa.aksw.org/gerbil/experiment?id=202605020011}{22.93} & \href{https://gerbil-qa.aksw.org/gerbil/experiment?id=202604160018}{21.79} & \href{https://gerbil-qa.aksw.org/gerbil/experiment?id=202605050107}{44.57} & \href{https://gerbil-qa.aksw.org/gerbil/experiment?id=202604270010}{\textbf{59.15}} & \href{https://gerbil-qa.aksw.org/gerbil/experiment?id=202605060005}{57.68} \\
\midrule
  & de & \href{https://gerbil-qa.aksw.org/gerbil/experiment?id=202605020003}{14.55} & \href{https://gerbil-qa.aksw.org/gerbil/experiment?id=202604160005}{23.11} & \href{https://gerbil-qa.aksw.org/gerbil/experiment?id=202605050109}{26.44} & \href{https://gerbil-qa.aksw.org/gerbil/experiment?id=202604270004}{59.34} & \href{https://gerbil-qa.aksw.org/gerbil/experiment?id=202605060011}{\textbf{62.02}} \\
\rowcolor{gray!20}
\cellcolor{white} & en & \href{https://gerbil-qa.aksw.org/gerbil/experiment?id=202605020001}{16.52} & \href{https://gerbil-qa.aksw.org/gerbil/experiment?id=202604160008}{29.57} & \href{https://gerbil-qa.aksw.org/gerbil/experiment?id=202605050111}{26.18} & \href{https://gerbil-qa.aksw.org/gerbil/experiment?id=202604270001}{62.29} & \href{https://gerbil-qa.aksw.org/gerbil/experiment?id=202605060009}{\textbf{66.29}} \\
  & ru & \href{https://gerbil-qa.aksw.org/gerbil/experiment?id=202605020006}{13.68} & \href{https://gerbil-qa.aksw.org/gerbil/experiment?id=202604160007}{27.30} & \href{https://gerbil-qa.aksw.org/gerbil/experiment?id=202605050110}{23.78} & \href{https://gerbil-qa.aksw.org/gerbil/experiment?id=202604270002}{52.40} & \href{https://gerbil-qa.aksw.org/gerbil/experiment?id=202605060010}{\textbf{58.49}} \\
\rowcolor{gray!20}
\cellcolor{white}\multirow{-4}{*}{\dataset{QALD-10}} & zh & \href{https://gerbil-qa.aksw.org/gerbil/experiment?id=202605020004}{13.30} & \href{https://gerbil-qa.aksw.org/gerbil/experiment?id=202604160009}{24.20} & \href{https://gerbil-qa.aksw.org/gerbil/experiment?id=202605050112}{21.00} & \href{https://gerbil-qa.aksw.org/gerbil/experiment?id=202604270000}{51.11} & \href{https://gerbil-qa.aksw.org/gerbil/experiment?id=202605060008}{\textbf{55.04}} \\
\midrule
\dataset{LC-QuAD2.0} & en & - & - & \href{https://gerbil-qa.aksw.org/gerbil/experiment?id=202604300000}{34.56} & \href{https://gerbil-qa.aksw.org/gerbil/experiment?id=202605080001}{38.36} & \href{https://gerbil-qa.aksw.org/gerbil/experiment?id=202605060014}{\textbf{40.83}} \\
\bottomrule
\end{tabular}
}%
\end{table}

\subsubsection{QALD-9-plus}
On this nine-language benchmark, \approach and \tool{GRASP} are well ahead of the other systems and split the per-language wins. \approach takes French, Bashkir, and Armenian; \tool{GRASP} takes English, German, Spanish, Russian, Belarusian, and Ukrainian. The large gaps in favor of \approach on French ($71.93$ vs.\ $12.38$) and Armenian ($65.96$ vs.\ $10.14$) reflect uneven language coverage in \dataset{QALD-9-plus}: only 23 French and 18 Armenian questions are available, against 127 for the fully covered languages, so per-language scores are sensitive to a handful of correct answers. \approach adapts to both high- and low-resource languages, whereas \tool{GRASP} performs better on fully covered languages, likely because of its optimizations for common KG patterns.

\subsubsection{QALD-10}
This dataset contains queries with aggregations and subqueries~\cite{usbeck2024qald}, and requires more hops than the other benchmarks in our work. 
\approach achieves the best F1 on all four languages in this dataset (English, German, Russian, and Chinese), outperforming every baseline, including \tool{GRASP}. Comprising roughly 30\% multi-hop and nested queries, in our evaluation, the dataset structure appears to align well with pipelines that explicitly plan traversal.

\subsubsection{LC-QuAD2.0}
Reproducing a large number of baselines is  computationally expensive, so we restricted the comparison to the strongest systems reported in prior benchmarks. 
\approach achieves the best F1, ahead of both \tool{UniQ-Gen} and \tool{GRASP}, without dataset-specific fine-tuning. \dataset{LC-QuAD2.0} is built from a fixed set of query templates, which usually favors fine-tuned methods like \tool{UniQ-Gen}.
\subsection{Error Analysis}\label{sec:error}

While \approach is evaluated across multiple languages, we focus our qualitative error analysis on the English subset, as these cases are representative of the general failure modes observed across the entire multilingual evaluation. We inspect the cases where \approach answers incorrectly on \dataset{QALD-9-plus} and \dataset{QALD-10}. Of the 99 answered questions in \dataset{QALD-9-plus}, 60 are correct; for \dataset{QALD-10}, 202 of 289. Of these correct answers, 26 and 95 respectively come from a SPARQL query that differs from the gold, showing that the system often finds an alternative formulation with the same answer set, a property SPARQL exact-match metrics miss. We refer to questions below by their dataset ID prefixed with \#. 

\subsubsection{Result Truncation}
\approach caps every generated query at 1{,}000 results to mitigate the risk of malformed queries returning excessively large result sets. 
For genuinely open-ended questions this may reduce recall: \dataset{QALD-9-plus} \#177 ``Which bridges are of the same type as the Manhattan Bridge?'' (gold $24{,}512$, pred $1{,}000$) and \#213 ``Show me all Czech movies'' (gold $8{,}376$, pred $1{,}000$) are representative examples. Some of these gold counts are likely inflated by Wikidata growth since the benchmark was released, but the cap remains the dominant factor. 

\subsubsection{Boolean Questions Answered with Evidence}
On \dataset{QALD-10}, the system often returns the underlying numeric or entity evidence instead of a yes/no answer. \#170 ``Do more than 100{,}000{,}000 people speak Japanese?'' returns $128{,}000{,}000$ rather than \texttt{true}, and \#223 ``Is Germany bigger than Poland?'' returns a comparison result instead of producing an \texttt{ASK} query. 

\subsubsection{Aggregation and Projection}
Comparative ``which is more Y, A or B?'' questions return both candidates and comparative values rather than only the winner (\dataset{QALD-10} \#361 ``who is older, Messi or Ronaldo?'', gold: \textit{wd:Q11571}, pred: \{\textit{(wd:Q11571, 1985-02-05), (wd:Q615, 1987-06-24)}\}), and some ``how many X'' questions are not able to return the correct value as the system does not consistently produce the expected structure. The system retrieves the right facts but does not consistently project them through \texttt{COUNT}. 

\subsubsection{Conservative or Imprecise Selection}
``When did X?'' and ``What year was X?'' questions often yield empty predictions because the generation step abstains rather than emit an incorrect literal (\dataset{QALD-10} \#238 ``What year did the Berlin Wall fall?'', gold $1989$), trading recall for precision. For broad ``Give me all X'' queries, the chosen class is occasionally too narrow: \dataset{QALD-9-plus} \#199 ``Give me all Argentine films'' returns one film against a gold set of $4{,}088$.
\newline
\newline
\noindent
Overall, \approach performs well across diverse benchmarks, languages, and query types without dataset-specific fine-tuning, with a clear advantage on the multi-hop and nested queries of \dataset{QALD-10}. The errors we observe, namely truncated results, missing \texttt{ASK} and \texttt{COUNT} projections, and overly cautious literal generation, can be resolved through targeted engineering and do not reflect deeper architectural flaws. 
These results suggest that fixed, planning-oriented pipelines may offer stronger generalization in our multi-hop questions setting than fine-tuned text-to-SPARQL models or agents that select tools at runtime. 
\section{Ablation Experiments}\label{sec:ablation}
We run ablation experiments on the \dataset{QALD-10}~\cite{usbeck2024qald} train set, supplying the gold entities (those extracted from the reference SPARQL query) during generation, 
Table~\ref{tab:model-features} lists evaluated features. 
A full factorial design is impractical, so we restrict the experiments to English queries and proceed in phases: each phase fixes one or two settings for the subsequent phases while still exploring the remaining feature combinations against those fixed values. In Table~\ref{tab:mars_ablation_all}, the configuration carried forward to the next phase appears in \textbf{bold}, and $^\dagger$ marks the locally best result in the two phases where it differs from the selection (Phases~1b and~4).

\begin{table}[t]
\caption{Overview of model features and settings.}
\label{tab:model-features}
\centering
\resizebox{\textwidth}{!}{%
\setlength{\tabcolsep}{4pt}
\begin{tabular}{lll}
\toprule
\textbf{Feature Name} & \textbf{Abbr.} & \textbf{Description} \\ \midrule
\rowcolor{gray!20} Model & \textsc{mod} & LLM used as \sparqlreasoner. \\
Top-N Similarity & \textsc{topn} & Number of highest-similarity patterns selected. \\
\rowcolor{gray!20} M-Hop Limit & \textsc{mhop} & Maximum hops from root entities for context gathering. \\
Augmented Text Similarity  & \textsc{aug} & Use entity-extraction augmented text for pattern similarity. \\
\rowcolor{gray!20} Concrete Examples & \textsc{conc} & Examples from the retrieved pattern instances in prompt. \\
Typological Information & \textsc{cls} & Utilize predicate domain/range in prompt. \\
\rowcolor{gray!20} Pattern Count & \textsc{pat} & Incorporate pattern frequency for each triple-pattern in prompt. \\
SPARQL Verification & \textsc{ver} & Second pass verifying and updating the generated SPARQL. \\ \bottomrule
\end{tabular}
}%
\end{table}
\begin{table}[!b]
\caption{Ablation results across all four phases (Macro F1 in [\%]).}
\label{tab:mars_ablation_all}
\centering
\small
\setlength{\tabcolsep}{5pt}
\setlength{\aboverulesep}{0pt}    
\setlength{\belowrulesep}{0pt}    
\renewcommand{\arraystretch}{1.25} 
\resizebox{\textwidth}{!}{%
\begin{tabularx}{\textwidth}{c l Y}
\toprule
\textbf{Phase} & \textbf{Decision} & \textbf{Top configurations (Macro F1)} \\
\midrule
\rowcolor{gray!20}
1a & \tool{GPT-OSS}
   & \textbf{\tool{GPT-OSS}\,@\,20/2: 65.77};\ \tool{Qwen3}\,@\,20/2: 63.81;\ \tool{Gemma3}\,@\,20/2: 58.26;\ \tool{GPT-OSS}\,@\,5/1: 56.59 \\
1b & \textsc{topn}/\textsc{mhop}\,=\,50/10
   & 500/1: 65.14$^\dagger$;\ \textbf{50/10: 63.50};\ 50/5: 63.50;\ 50/1: 63.24;\ 100/2: 63.17 \\
\rowcolor{gray!20}
2  & \textsc{conc}
   & \textbf{\textsc{conc}: 65.47};\ \textsc{cls}: 65.07;\ \textsc{aug},\,\textsc{conc}: 64.97;\ \textsc{aug},\,\textsc{cls},\,\textsc{conc}: 63.97;\ \textsc{aug},\,\textsc{cls}: 61.05 \\
3  & \textsc{ver},\,\textsc{conc}
   & \textbf{\textsc{ver},\,\textsc{conc}: 70.94};\ \textsc{pat},\,\textsc{ver},\,\textsc{conc}: 70.30;\ \textsc{pat},\,\textsc{conc}: 67.14;\ \textsc{conc}: 65.47 \\
\rowcolor{gray!20}
4  & \textsc{topn}/\textsc{mhop}\,=\,20/10
   & 100/1: 72.10$^\dagger$;\ 50/10: 68.67;\ 20/1: 68.61;\ 50/1: 68.47;\ 20/5: 68.03;\ \textbf{20/10: 67.92};\ 100/10: 66.02 \\
\bottomrule
\end{tabularx}
}
\end{table}
\subsection{Phase 1: Model Selection and Retrieval Grid}\label{subsec_abl_p1}
We evaluate three open-weight LLMs as $\operatorname{\sparqlreasoner}$ (\tool{Qwen3}~\cite{qwen3}, \tool{Gemma3}, \tool{GPT-OSS}~\cite{openai2025gptoss120bgptoss20bmodel}) at two operating points (\textsc{topn}/\textsc{mhop}\,=\,5/1, 20/2), then sweep \textsc{topn}\,$\in$\,\{5,10,20,50,100,500\} against \textsc{mhop}\,$\in$\,\{1,2,5,10,20,50\} for the top-scoring model. \tool{GPT-OSS} leads at both operating points and, being a sparse Mixture-of-Experts model, runs at lower inference cost than comparable dense models, so we fix \textsc{mod}=\tool{GPT-OSS} for the remaining phases. The largest grid cell (\textsc{topn}=500, \textsc{mhop}=1) reaches the highest F1 ($65.14$) but exceeds the \sparqlreasoner's context window once additional features are enabled; we therefore use \textsc{topn}=50, \textsc{mhop}=10 for the feature ablations, which sits within 2 F1 points of the unconstrained maximum and remains tractable.

\subsection{Phase 2: Augmented Text, Schema, and Examples}\label{subsec_abl_p2}
A $2^3$ factorial analysis over \textsc{aug}, \textsc{cls}, and \textsc{conc} at the Phase~1 operating point shows that only configurations with \textsc{conc} active beat the no-feature baseline ($63.50$). \textsc{conc} alone yields the largest single-feature gain; pairwise and three-way combinations with \textsc{aug} and \textsc{cls} all degrade performance, which we attribute to context dilution at \textsc{topn}=50, where the prompt is already information-dense. We carry \textsc{conc} forward to Phase~3.

\subsection{Phase 3: Pattern Frequency and Verification}\label{subsec_abl_p3}
A $2^2$ factorial analysis over \textsc{pat} and \textsc{ver} on top of the Phase~2 setting shows that \textsc{ver} yields a sizeable gain (from $65.47$ to $70.94$), \textsc{pat} alone is neutral-to-slightly-negative, and their combination falls between the two. Verification corrects systematic SPARQL errors from the first generation pass.

\subsection{Phase 4: Full-Feature Scaling}\label{subsec_abl_p4}
We enable the full feature set (\textsc{aug}+\textsc{cls}+\textsc{conc}+\textsc{pat}+\textsc{ver}) and re-sweep \textsc{topn} against \textsc{mhop} to check whether the retrieval--depth trade-off shifts. Two findings stand out. \textsc{topn}=100 with \textsc{mhop}=1 gives the highest train-set F1 ($72.10$) but is the most expensive (avg.\ $12.3$\,k tokens per request) and relies on shallow retrieval, which is unlikely to generalise to multi-hop queries. Raising \textsc{mhop} beyond 10 has no measurable effect (\textsc{topn}=50 yields the same F1 at \textsc{mhop}=10 and \textsc{mhop}=20), confirming that the relevant subgraph is fully covered before the limit.

\subsection{Final Configuration}
We adopt \tool{GPT-OSS} with \textsc{topn}=20, \textsc{mhop}=10, and all features enabled. We use \textsc{topn}=20 rather than 100 despite the latter's higher train-set F1 ($72.10$ vs.\ $67.92$): the 20-pattern setting consumes a third of the tokens per request ($4.3$\,k vs.\ $12.3$\,k) at a cost of about 4 F1 points on the train set, which we prefer to keep test-time inference cost bounded and to avoid overfitting. We also keep \textsc{aug} and \textsc{cls} enabled despite their negative isolated contribution in Phase~2, since both inject grounding signals (entity-augmented text, predicate domain/range) that the model can fall back on when surface forms diverge from the schema, which we expect on the multilingual test splits where lexical overlap with the English-trained pattern index is weaker.

As a pipeline sanity check, we compare against \textsc{SimpleSparqlGenerator} (\tool{SSG}), which asks \tool{GPT-OSS} to generate SPARQL directly from the same entity and relation links. \tool{SSG} reaches Macro F1 $52.34$ against $67.92$ for our final configuration, isolating the pipeline's contribution beyond the underlying LLM.

%
%
%
\section{Conclusion and Limitations}\label{sec:Conclusion}
In this work, we introduced \approach, a retrieval and reasoning pipeline for KGQA over Wikidata that operates without dataset-specific fine-tuning. The system combines iterative guided context enrichment with the KG schema and instance facts, and supports multilingual input through augmented translation. Across the three benchmarks, \approach achieves the best Macro F1 on \dataset{QALD-10} (all four languages) and on \dataset{LC-QuAD2.0}, and splits the per-language wins with the strongest baseline (\tool{GRASP}) on \dataset{QALD-9-plus}. The advantage of MARS is most evident on multi-hop and aggregation-heavy queries, consistent with our use of an explicit pipeline rather than a tool-using agent. The shortcomings are primarily centered around technical constraints and output errors identified in Section~\ref{sec:error}, pointing toward concrete directions for further improvement. \approach is built on open-weight LLMs and the code is publicly available to support reproducibility.
\approach{'s} limitations fall into two groups: those internal to the system, and those inherited from the current state of KGQA evaluation. For the latter, we highlight contributions that aim to address them for any future system. 

\textit{F1 is not directly comparable across papers.} KGQA systems report Macro F1 with different denominators: some over predicted answer sets, others over SPARQL exact-match, others over SPARQL components such as entities and relations. Headline numbers from other papers therefore cannot be ranked directly. To ensure consistent comparisons, we re-run every baseline through the same answer-set protocol rather than quoting numbers from respective papers.

\textit{Benchmark--KG version drift.} 
As noted in Section \ref{sec:intro}, benchmarks often lack the specific KG snapshots used during construction, as seen with \dataset{QALD-10}'s incomplete Wikidata dump. To ensure reproducibility, we release the exact Wikidata snapshot used alongside our updated datasets.

\textit{Pattern retrieval on high-degree nodes.} Pattern retrieval over high-degree nodes in Wikidata-scale KGs requires joins that most triple-stores cannot serve in reasonable time. We rely on \tool{Tentris} to make this tractable, while broader triple-store compatibility remains an engineering target.

\textit{Benchmark coverage.} Resource constraints limited our evaluation to Wikidata-based benchmarks; extending \approach to other KGs such as Freebase and DBpedia is future work. We also attempted to include \dataset{SPINACH}~\cite{liu-etal-2024-spinach} for its higher query complexity, but its gold queries depend on Wikidata-specific services such as \texttt{wikibase:label} that cannot be deployed locally. Rewriting them to use \texttt{rdfs:label} partially mitigated the issue but introduced others, and \dataset{SPINACH}'s answer-set conventions differ enough from \dataset{QALD} and \dataset{LC-QuAD2.0} that a fair evaluation would require additional prompt engineering.

\section*{Acknowledgements}
This work has been funded by the Deutsche Forschungsgemeinschaft (DFG, German Research Foundation) – TRR 318/3 2026 – 438445824, by the Federal Ministry for Economic Affairs and Energy (BMWE) on the basis of a decision by the German Bundestag under the project ikDS (FKZ KK5175206LO4), by the German Federal Ministry of Research, Technology and Space (BMFTR) within the projects Learn2RAG (01MK250104) and KI-Akademie OWL (16IS24057B), and by the Ministry of Culture and Science of North Rhine-Westphalia (MKW NRW) within the project SAIL (NW21-059D).

We would like to thank Lennart Austenfeld for his assistance with the configuration and deployment of the large language models.

Generative AI tools (ChatGPT, Claude, Gemma, Qwen) were used for language refinement and code assistance. All scientific contributions and conclusions are solely those of the authors, who reviewed and take full responsibility for the final manuscript.

%
%
%
% ...

\bibliographystyle{splncs04}
\bibliography{merged}


\end{document}